\documentclass{article}

\usepackage[final]{tackling_climate_workshop_style}

\usepackage[utf8]{inputenc} 
\usepackage[T1]{fontenc}    
\usepackage{hyperref}       
\usepackage{url}            
\usepackage{booktabs}       
\usepackage{amsfonts}       
\usepackage{nicefrac}       
\usepackage{microtype}      
\usepackage{amsmath}
\usepackage[normalem]{ulem}
\usepackage{xcolor}
\usepackage{graphicx}

\title{Probabilistic bias adjustment of seasonal predictions of Arctic Sea Ice Concentration}

\author{%
  Parsa Gooya \\
  Canadian Centre for Climate Modleing and Analysis, Environment and Climate Change Canada\\
  Victoria, British Columbia, Canada \\
  \texttt{parsa.gooya@ec.gc.ca} \\
  \And
  Reinel Sospedra-Alfonso \\
  Canadian Centre for Climate Modleing and Analysis, Environment and Climate Change Canada \\
  Victoria, British Columbia, Canada  \\
  \texttt{reinel.sospedra-alfonso@ec.gc.ca} 
}

\begin{document}

\maketitle

\begin{abstract}
Seasonal forecast of Arctic sea ice concentration is key to mitigate the negative impact and assess potential opportunities posed by the rapid decline of sea ice coverage. Seasonal prediction systems based on climate models often show systematic biases and complex spatio-temporal errors that grow with the forecasts. Consequently, operational predictions are routinely bias corrected and calibrated using retrospective forecasts. For predictions of Arctic sea ice concentration, error corrections are mainly based on one-to-one post-processing methods including climatological mean or linear regression correction and, more recently, machine learning. Such deterministic adjustments are confined at best to the limited number of costly-to-run ensemble members of the raw forecast. However, decision-making requires proper quantification of uncertainty and likelihood of events, particularly of extremes. We introduce a probabilistic error correction framework based on a conditional Variational Autoencoder model to map the conditional distribution of observations given the biased model prediction. This method naturally allows for  generating large ensembles of adjusted forecasts. We evaluate our model using deterministic and probabilistic metrics and show that the adjusted forecasts are better calibrated, closer to the observational distribution, and have smaller errors than climatological mean adjusted forecasts.  
\end{abstract}

\section{Introduction}

Satellite observations indicate rapid decline in Arctic sea ice concentration and extent across all calender months \cite{serreze2007, Cavalieri_Parkinson_2012}. The decline in Arctic sea ice, especially in summer, poses a threat to local communities and ecosystems, but also creates economic opportunities for marine fishing, shipping, tourism and resource
extraction \cite{Sea_Ice_Initialisation_On_Seasonal_Prediction_Skill, SubseasonaltoSeasonalArcticSeaIceForecastSkillImprovementfromSeaIceConcentrationAssimilation}. Preparing, mitigating and planning accordingly in response to such changes and their impacts demands accurate and reliable seasonal predictions of Arctic sea ice \cite{Wagner02072020}.


Seasonal predictions refer to forecasts on time scales ranging from a few months to (slightly longer than) a year. The Canadian Seasonal to Interannual Prediction System \cite[CanSIPS,][]{canSIPS_2013, CanSIPsv2_2020} provides seasonal predictions of key climate variables including Arctic Sea Ice Concentration (SIC) using two coupled climate models. For CanSIPSv3, one such model is the Canadian Earth System Model version 5 \cite[CanESM5,][]{Swart-2019} developed at the Canadian Centre for Climate Modelling and Analysis. These seasonal forecasts are 12-month model simulations initialized each month using observation-based estimates of the climate system's current state. Various versions of CanSIPS have demonstrated notable skill in forecasting Arctic sea ice \cite{Martin2023, Dirkson2021, sigmond2016, sigmond2013}. Nonetheless, inherent model deficiencies give rise to complex (often non-linear) systematic errors that increase with lead time, defined as the number of months ahead of the initialization date for which a forecast is issued. As a result, post-processing techniques for error correction are routinely employed to reduce such biases.

For predictions of Arctic SIC, often the relatively simple lead time dependent climatological mean error correction or linear regression adjustments are used \cite{Blanchard-Wrigglesworth2017}. More recently, Palerme et al. (2024) \cite{Palerme2024} and He et al. (2025) \cite{He2025MLICE} applied machine learning (ML) to improve the skill of sea ice concentration forecasts on weather and seasonal time scales of Arctic sea ice. However such ML-based bias corrections, rely mainly on deterministic models for adjustment and are at best confined to the limited number of costly-to-run prediction ensemble members of the original model simulations. Uncertainty quantification is key to reliable seasonal forecasts and decision making relies on (calibrated) probabilistic forecasts that would sample different weather events (e.g., extremes) and allow the estimation of their likelihood. Here, we introduce a probabilistic bias correction scheme based on a conditional generative ML model. A conditional Variational Autoencoder (cVAE) is developed and applied to adjust seasonal predictions of Arctic SIC produced with CanESM5 and contributing to CanSIPSv3.
Different approaches to uncertainty quantification in ML-based corrections of sub-seasonal to decadal predictions have been proposed in previous studies \cite{Sospedra_Alfonso_2024, sacco_2022, Gromquist_DLpost-processing, diffensemblegen2024}. However, we are not aware of any previous studies using generative machine learning for probabilistic bias adjustment (one-to-many) of model-based seasonal forecasts.

\section{Data and Methods}
\subsection{Generative Framework}
Introduced by Sohn et al. (2015) \cite{cVAE2015}, conditional VAEs (cVAEs) are modifications of VAEs \cite{kingma2014, rezende2014}intended to learn the conditional distribution of data by modeling the data variable ${x}$ with the help of an unobservable (latent) variable ${z}$ \cite{cVAE2015, prince2023understanding}, conditional on $c$. The latent variable $z$ can be regarded as a lower-dimensional representation of ${x}$, coming from the conditional prior distribution $p({z|c})$ that explains the variations in data in a simpler manner. The generative process uses a sample of latent variables ${z} \in \mathcal{Z}$ from the conditional prior distribution $p({z|c})$ which is generally assumed to be a Gaussian; a normal distribution whose parameters depend on the condition. The data ${x} \in \mathcal{X}$ is generated (sampled) from the distribution $p_{\theta}({x} | z, c)$, also generally parametrized as a Gaussian \cite{cVAE2015}. The parameters of this generative distribution $p_{\theta}({x} | z, c)$ are estimated using a neural network, referred to as the probabilistic $\textit{decoder}$. Distribution parameters are learned using maximum likelihood estimation, which optimizes the parameters  $\theta$ of the neural network model to maximize the likelihood of the generated samples \cite{cVAE2015}. Kingma \& Welling (2014) showed that the parameters of the VAE can be estimated efficiently using the variational lower bound of the log-likelihood as a surrogate objective function. For the cVAE the surrogate objective function is:

\begin{equation}
\begin{split}
\log p(x|c) =  KL \left( q_{\phi}({z} | x,c) \Vert p({z} | x, c) \right) + \mathbb{E}_{q_{\phi}({z} | x, c)} \left[ - \log q_{\phi}({z} | x, c) + \log p({x}, {z} | c) \right]\\ 
\geq - KL \left( q_{\phi}({z} | x, c) \Vert p(z|c) \right) + \mathbb{E}_{q_{\phi}({z} | x, c)} \left[ \log p_{\theta}({x} | z, c) \right],
\end{split}
\end{equation}

\noindent where the term $KL$ refers to the Kullback–Leibler Divergence. In this framework, the distribution $q_{\phi}({z} | x, c)$ (also assumed to be a Gaussian), is introduced to approximate the true intractable posterior $p({z} | x, c)$ \cite{kingma2014, cVAE2015} and is parametrized using another neural network with parameters $\phi$ --the probabilistic \textit{encoder}. 

For probabilistic bias correction, our goal is to learn a probabilistic mapping from biased \textit{ensemble mean} model predictions ($\bar x_{tl}$ where $t$ stands for initialization time and $l$ for lead time) of SIC to the observational distribution $p(Y|\bar x_{tl})$. Thus, we maximize the likelihood of the observation $y_{tl}$ conditioned on $\bar x_{tl}$ ($\max{_\theta}$  $p_{\theta}(Y = y_{tl}| \bar x_{tl})$). It is worth highlighting that the observation $y_{tl}$ is only a realization of the target observational distribution which we are modeling using the cVAE model. The cVAE model is thus formulated as follows:

\begin{equation}
    \begin{aligned}
        &\textit{Encoder:} \quad q_{\phi}({z} | {y_{tl}}, {\bar x_{tl}}) = \mathcal{N} \left( {\mu}_{NN_{\phi}}({y_{tl}}, {\bar x_{tl}}), {\sigma}_{NN_{\phi}}^2({y_{tl}}, {\bar x_{tl}}) {I} \right) \\
        &\textit{Decoder:} \quad p_{\theta}({y} | {z}, {\bar x_{tl}}) = \mathcal{N} \left( {\mu}_{NN_{\theta}}({z}, {\bar x_{tl}}), {\sigma^2}{I} \right) \\
        &\textit{Prior:} \quad p({z} | {\bar x_{tl}}) = \mathcal{N} \left( {\mu}_{NN_{\omega}}({\bar x_{tl}}), {\sigma}_{NN_{\omega}}^2( {\bar x_{tl}}) {I} \right)\\
    \end{aligned}
\end{equation}

\noindent where ${I}$ is the identity matrix and ${\sigma^2}$ is a location independent (constant) decoder noise. The parameters of the conditional prior distribution are learned using another neural network with parameters $\omega$. Under this formulation, the second term in the VAE loss function (Eq. 1) is proportional to the Mean Square Error (MSE) of the generated samples and the KL term has a closed form solution \cite{cVAE2015}.

\subsection{Architecture and Inference}

The architecture used for this study follows that of Sohn et al. (2015) with few modifications. The encoder and prior networks are series of double convolution blocks using a variation of ConvNeXt blocks \cite{Samudra, ConvNext2020} with partial convolution layers \cite{partialconv}, followed by maxpooling downsampling, mapping to a 1000-dimensional latent space. The decoder reverses the operations in the encoder (prior) networks using upsampling and double ConvNeXt blocks, mapping the latent samples back to the data space. The details of the architecture and training can be found in Appendix A.1. At inference, samples from the prior distribution are passed to the decoder. While the reverse KL divergence term in the loss penalizes divergence of the latent structure from the prior distribution, the latent encodings often diverge from falling perfectly under the prior distribution \cite{distVAE2023}. The normal prior in the cVAE enables controlling data synthesis by defining suitable loci in the distribution for sampling \cite{IEEE}. Following \cite{Gooya_2025, IEEE}, we find a scaling factor for the prior standard deviation based on the spread over error ratio metric (SOE, Sec. 3) over the validation period. The scaling factor allows sampling a wider range of internal variability at inference time and a more reliable forecast (see Appendix A.3).

\subsection{Data}

We post-process retrospective seasonal forecasts of Arctic SIC produced with CanSIPSv3's CanESM5 model. Each forecast consists of 10 ensemble members of 12-months predictions and is initialized at the beginning of each month starting in January 1980. The predictions are remapped to the standard $1\times1$ grid covering latitudes above $50^{\circ}$ North. We use retrospective predictions up to December 2015 as training data and reserve the forecasts issued in January 2016 to December 2018 as a validation set. The adjusted results are tested for the years 2019 to 2020. We employ a temporal mask \cite{Sospedra_Alfonso_2024} to ensure that no future data leak during training from the validation set, and from the test set to the validation set. The target observational data are from the satellite-based NOAA/NSIDC Climate Data Record of passive microwave SIC v4 \cite{SIC_NASA_CDR_v4} spanning January 1981 to December 2021. These data are remapped to the same grid and location as above, and re-arranged into a structure similar to the model data consisting of a monthly forecasts each spanning over 12 months from the initialization date. 

\section{Evaluation metrics}

We compare the cVAE corrected ensemble (nadj, 100 members) with a lead time dependent climatological mean adjusted ensemble forecast as benchmark (badj, 10 members) \cite{Sospedra_Alfonso_2024}. In a well-calibrated forecast ensemble, the verification data should be indistinguishable from any member of the ensemble. This is usually expressed using rank histograms, which measures the distribution of the rank of the observed field (obs) in the forecast ensemble. We measure the rank at each grid point and report Cumulative Distribution Functions (CDF) of rank histograms \cite{daust2024capturing}. For a calibrated system, each rank should have the same probability of occurrence, so the histogram should estimate a uniform distribution, corresponding to a the 1:1 line CDF. If the CDF has more weight at the tails (a U-shaped histogram), the ensemble forecast is overconfident. The noise variance and mean square error (MSE) define the hindcast SOE, which measures the reliability of the ensemble. SOE = 1 indicates that the ensemble members and observations are statistically indistinguishable \cite{SOE2013}, whereas SOE < 1 (SOE > 1) indicates over(under) confidence. Finally, the corrected ensemble mean is compared to the observation using RMSE at grid cell level, RMSE of integrated measures of sea ice area (SIA) and extent (SIE), as well as using Integrated Ice Edge Error (IIEE) which measures the difference between the areas enclosed by the predicted and true ice edges \cite{IIEE}. See Appendix A.2 for definitions.

\section{Results and Discussions}



\begin{figure}
\centering
\includegraphics[width=\textwidth]{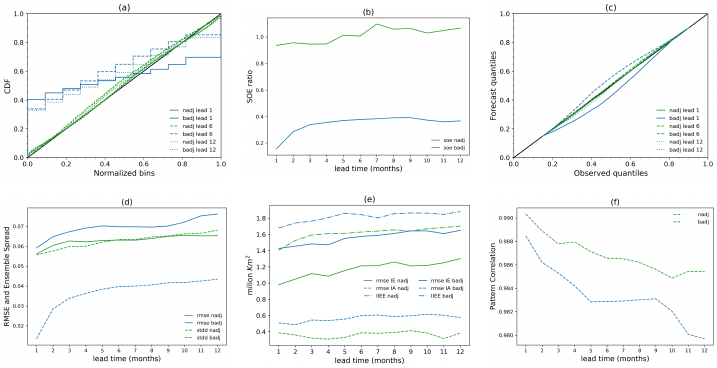}
\caption{a) CDF of rank histograms of the nadj/badj versus lead times measured at marginal ice grid cells. Only three lead times are plotted for visibility. b) SOE versus lead time showing reliability. c) QQ plots at three lead times comparing the distribution of SIC at marginal ice grid cells with obs. d) RMSE (solid) over initialization time between the ensemble mean nadj/badj compared to obs at grid cells level averaged over the entire region. The dashed line shows the global mean ensemble spread averaged over initialization time. e) For each lead time, RMSE of SIA (solid line) and SIE (dashed line) over initialization time, and average IIEE (dotted line) over intialization time is compared between ensemble mean nadj/badj and obs. f) same as (e) but for pattern correlation relative to obs.} 
\label{fig:fig1}
\end{figure}

Fig. 1a-c compare the probabilistic performance of corrected ensembles relative to the observation. CDF curves for nadj (100 members) show close to uniform rank histograms indicating that observations have similar probability of falling at every rank, confirming a well-calibrated ensemble. This is further validated using SOE ratio, which is consistently close to 1 for nadj, indicating that the ensemble members are indistinguishable from observation. The limited badj ensemble (10 members) predictions are clearly overconfident with a heavy-tailed CDF and SOE<1. Finally, the QQ plots confirm that the distribution of the SIC in the corrected ensemble remains close to the observational distribution. The CDF and QQ plots are reported at critical marginal ice grid cells (defined as $0.15\leq \hbox{SIC} \leq0.90$) to avoid being biased by performance over fully covered or open ocean regions. Comparing ensemble mean of nadj and badj with observations using deterministic metrics (Fig. 1d-f) confirms that both at grid level and for bulk measures of integrated ice coverage (SIA), extent (SIE) and importantly, the boundaries of the edges of ice (IIEE), the nadj ensemble mean is more accurate than badj. The error increase with lead time is expected for all operational forecasts as predicting further in time is more uncertain. This is reflected in the increase of ensemble variance (Fig. 1d). However, Fig. 1b shows that while longer predictions are less confident, the SOE ratio remains consistently close to 1 for nadj. Moreover, Fig. 1a shows that while the reliability decreases as expected, the CDFs for nadj remain fairly close to the 1:1 line. Possible caveats include spectral bias (blurry images), under-representation of decoder noise \cite{Gooya_2025, dorta2018structureduncertaintypredictionnetworks}, and understanding the source of the generated variability which will be addressed in a future study. Additionally, in the current setting, the generated ensemble members are independent in time dimension. To address that, future research will explore running the model autoregressively to generate corrected ensemble members that are traceable in time. These results confirm that probabilistic forecast adjustment based on cVAE  provide fast, skillful and reliable arbitrary large ensembles of corrected forecast. 

\newpage

\bibliographystyle{unsrt}
\bibliography{References}

\begin{thebibliography}{10}

\bibitem{serreze2007}
Mark~C. Serreze, Marika~M. Holland, and Julienne Stroeve.
\newblock Perspectives on the arctic's shrinking sea-ice cover.
\newblock {\em Science}, 315(5818):1533--1536, 2007.

\bibitem{Cavalieri_Parkinson_2012}
D.~J. Cavalieri and C.~L. Parkinson.
\newblock Arctic sea ice variability and trends, 1979--2010.
\newblock {\em The Cryosphere}, 6(4):881--889, 2012.

\bibitem{Sea_Ice_Initialisation_On_Seasonal_Prediction_Skill}
M.~Kimmritz, F.~Counillon, L.~H. Smedsrud, I.~Bethke, N.~Keenlyside, F.~Ogawa, and Y.~Wang.
\newblock Impact of ocean and sea ice initialisation on seasonal prediction skill in the arctic.
\newblock {\em Journal of Advances in Modeling Earth Systems}, 11(12):4147--4166, 2019.

\bibitem{SubseasonaltoSeasonalArcticSeaIceForecastSkillImprovementfromSeaIceConcentrationAssimilation}
Yong-Fei Zhang, Mitchell Bushuk, Michael Winton, Bill Hurlin, Thomas Delworth, Matthew Harrison, Liwei Jia, Feiyu Lu, Anthony Rosati, and Xiaosong Yang.
\newblock Subseasonal-to-seasonal arctic sea ice forecast skill improvement from sea ice concentration assimilation.
\newblock {\em Journal of Climate}, 35(13):4233 -- 4252, 2022.

\bibitem{Wagner02072020}
Penelope~Mae Wagner, Nick Hughes, Pascale Bourbonnais, Julienne Stroeve, Lasse Rabenstein, Uma Bhatt, Joe Little, Helen Wiggins, and Andrew Fleming.
\newblock Sea-ice information and forecast needs for industry maritime stakeholders.
\newblock {\em Polar Geography}, 43(2-3):160--187, 2020.

\bibitem{canSIPS_2013}
William~J. Merryfield, Woo-Sung Lee, George~J. Boer, Viatcheslav~V. Kharin, John~F. Scinocca, Gregory~M. Flato, R.~S. Ajayamohan, John~C. Fyfe, Youmin Tang, and Saroja Polavarapu.
\newblock The canadian seasonal to interannual prediction system. part i: Models and initialization.
\newblock {\em Monthly Weather Review}, 141(8):2910 -- 2945, 2013.

\bibitem{CanSIPsv2_2020}
Hai Lin, William~J. Merryfield, Ryan Muncaster, Gregory~C. Smith, Marko Markovic, Frédéric Dupont, François Roy, Jean-François Lemieux, Arlan Dirkson, Viatcheslav~V. Kharin, Woo-Sung Lee, Martin Charron, and Amin Erfani.
\newblock The canadian seasonal to interannual prediction system version 2 (cansipsv2).
\newblock {\em Weather and Forecasting}, 35(4):1317 -- 1343, 2020.

\bibitem{Swart-2019}
N.~C. Swart, J.~N.~S. Cole, V.~V. Kharin, M.~Lazare, J.~F. Scinocca, N.~P. Gillett, J.~Anstey, V.~Arora, J.~R. Christian, S.~Hanna, Y.~Jiao, W.~G. Lee, F.~Majaess, O.~A. Saenko, C.~Seiler, C.~Seinen, A.~Shao, M.~Sigmond, L.~Solheim, K.~von Salzen, D.~Yang, and B.~Winter.
\newblock The canadian earth system model version 5 (canesm5.0.3).
\newblock {\em Geoscientific Model Development}, 12(11):4823--4873, 2019.

\bibitem{Martin2023}
Joseph Martin, Adam Monahan, and Michael Sigmond.
\newblock Improved seasonal forecast skill of pan-arctic and regional sea ice extent in cansips version 2.
\newblock {\em Weather and Forecasting}, 38(10):2029 -- 2056, 2023.

\bibitem{Dirkson2021}
A.~Dirkson, B.~Denis, M.~Sigmond, and W.~J. Merryfield.
\newblock Development and calibration of seasonal probabilistic forecasts of ice-free dates and freeze-up dates.
\newblock {\em Weather and Forecasting}, 36:301--324, 2021.

\bibitem{sigmond2016}
M.~Sigmond, M.~C. Reader, G.~M. Flato, W.~J. Merryfield, and A.~Tivy.
\newblock Skillful seasonal forecasts of arctic sea ice retreat and advance dates in a dynamical forecast system.
\newblock {\em Geophysical Research Letters}, 43(24):12,457--12,465, 2016.

\bibitem{sigmond2013}
M.~Sigmond, J.~C. Fyfe, G.~M. Flato, V.~V. Kharin, and W.~J. Merryfield.
\newblock Seasonal forecast skill of arctic sea ice area in a dynamical forecast system.
\newblock {\em Geophysical Research Letters}, 40(3):529--534, 2013.

\bibitem{Blanchard-Wrigglesworth2017}
E.~Blanchard-Wrigglesworth, A.~Barthélemy, M.~Chevallier, R.~Cullather, N.~Fučkar, F.~Massonnet, P.~Posey, W.~Wang, J.~Zhang, C.~Ardilouze, C.~M. Bitz, G.~Vernieres, A.~Wallcraft, and M.~Wang.
\newblock Multi-model seasonal forecast of arctic sea-ice: forecast uncertainty at pan-arctic and regional scales.
\newblock {\em Climate Dynamics}, 49(4):1399--1410, 2017.

\bibitem{Palerme2024}
C.~Palerme, T.~Lavergne, J.~Rusin, A.~Melsom, J.~Brajard, A.~F. Kvanum, A.~Macdonald~S{\o}rensen, L.~Bertino, and M.~M\"uller.
\newblock Improving short-term sea ice concentration forecasts using deep learning.
\newblock {\em The Cryosphere}, 18(4):2161--2176, 2024.

\bibitem{He2025MLICE}
Z.~He, Y.~Wang, J.~Brajard, X.~Wang, and Z.~Shen.
\newblock Improving seasonal arctic sea ice predictions with the combination of machine learning and earth system model.
\newblock {\em EGUsphere [preprint]}, 2025.

\bibitem{Sospedra_Alfonso_2024}
Reinel Sospedra-Alfonso, Parsa Gooya, and Johannes Exenberger.
\newblock Adjustment of decadal ocean carbon sink predictions using deep learning.
\newblock October 2024.

\bibitem{sacco_2022}
Maximiliano Sacco, Juan Ruiz, Manuel Pulido, and Pierre Tandeo.
\newblock {Evaluation of Machine Learning Techniques for Forecast Uncertainty Quantification}.
\newblock {\em {Quarterly Journal of the Royal Meteorological Society}}, 148(749):3470--3490, August 2022.

\bibitem{Gromquist_DLpost-processing}
Peter Grönquist, Chengyuan Yao, Tal Ben-Nun, Nikoli Dryden, Peter Dueben, Shigang Li, and Torsten Hoefler.
\newblock Deep learning for post-processing ensemble weather forecasts.
\newblock {\em Philosophical Transactions of the Royal Society A: Mathematical, Physical and Engineering Sciences}, 379(2194):20200092, 2021.

\bibitem{diffensemblegen2024}
Lizao Li, Robert Carver, Ignacio Lopez-Gomez, Fei Sha, and John Anderson.
\newblock Generative emulation of weather forecast ensembles with diffusion models.
\newblock {\em Science Advances}, 10(13):eadk4489, 2024.

\bibitem{cVAE2015}
Kihyuk Sohn, Honglak Lee, and Xinchen Yan.
\newblock Learning structured output representation using deep conditional generative models.
\newblock 28, 2015.

\bibitem{kingma2014}
Diederik~P. Kingma and Max Welling.
\newblock Auto-encoding variational bayes.
\newblock 2014.

\bibitem{rezende2014}
Danilo~Jimenez Rezende, Shakir Mohamed, and Daan Wierstra.
\newblock Stochastic backpropagation and approximate inference in deep generative models.
\newblock 32(2):1278--1286, 22--24 Jun 2014.

\bibitem{prince2023understanding}
Simon~J.D. Prince.
\newblock {\em Understanding Deep Learning}.
\newblock The MIT Press, 2023.

\bibitem{Samudra}
Surya Dheeshjith, Adam Subel, Alistair Adcroft, Julius Busecke, Carlos Fernandez-Granda, Shubham Gupta, and Laure Zanna.
\newblock Samudra: An ai global ocean emulator for climate.
\newblock {\em Geophysical Research Letters}, 52(10):e2024GL114318, 2025.
\newblock e2024GL114318 2024GL114318.

\bibitem{ConvNext2020}
Zhuang Liu, Hanzi Mao, Chao-Yuan Wu, Christoph Feichtenhofer, Trevor Darrell, and Saining Xie.
\newblock A convnet for the 2020s.
\newblock In {\em 2022 IEEE/CVF Conference on Computer Vision and Pattern Recognition (CVPR)}, pages 11966--11976, 2022.

\bibitem{partialconv}
Guilin Liu, Fitsum~A. Reda, Kevin~J. Shih, Ting-Chun Wang, Andrew Tao, and Bryan Catanzaro.
\newblock Image inpainting for irregular holes using partial convolutions, 2018.

\bibitem{distVAE2023}
Seunghwan An and Jong-June Jeon.
\newblock Distributional learning of variational autoencoder: Application to synthetic data generation.
\newblock 36:57825--57851, 2023.

\bibitem{IEEE}
Dario A.~B. Oliveira, Jorge~G. Diaz, Bianca Zadrozny, Campbell~D. Watson, and Xiao~Xiang Zhu.
\newblock Controlling weather field synthesis using variational autoencoders.
\newblock pages 5027--5030, 2022.

\bibitem{Gooya_2025}
Parsa Gooya, Reinel Sospedra-Alfonso, and Johannes Exenberger.
\newblock Toward generative machine learning for boosting ensembles of climate simulations.
\newblock July 2025.

\bibitem{SIC_NASA_CDR_v4}
Walt Meier, F.~Fetterer, A.~Windnagel, and S.~Stewart.
\newblock Noaa/nsidc climate data record of passive microwave sea ice concentration, version 4, 2021.

\bibitem{daust2024capturing}
Kiri Daust and Adam Monahan.
\newblock Capturing climatic variability: Using deep learning for stochastic downscaling.
\newblock {\em arXiv preprint arXiv:2406.02587}, 2024.
\newblock Submitted to Artificial Intelligence for the Earth Systems AMS Journal.

\bibitem{SOE2013}
Chun~Kit Ho, Ed~Hawkins, Len Shaffrey, Jochen Bröcker, Leon Hermanson, James~M. Murphy, Doug~M. Smith, and Rosie Eade.
\newblock Examining reliability of seasonal to decadal sea surface temperature forecasts: The role of ensemble dispersion.
\newblock {\em Geophysical Research Letters}, 40(21):5770--5775, 2013.

\bibitem{IIEE}
H.~F. Goessling, S.~Tietsche, J.~J. Day, E.~Hawkins, and T.~Jung.
\newblock Predictability of the arctic sea ice edge.
\newblock {\em Geophysical Research Letters}, 43(4):1642--1650, 2016.

\bibitem{dorta2018structureduncertaintypredictionnetworks}
Garoe Dorta, Sara Vicente, Lourdes Agapito, Neill D.~F. Campbell, and Ivor Simpson.
\newblock Structured uncertainty prediction networks, 2018.

\bibitem{Odena2016}
A.~Odena, V.~Dumoulin, and C.~Olah.
\newblock Deconvolution and checkerboard artifacts.
\newblock {\em Distill}, 1(10):e3, October 2016.

\bibitem{finn2024diffice}
Tobias~Sebastian Finn, Charlotte Durand, Alban Farchi, Marc Bocquet, and Julien Brajard.
\newblock Towards diffusion models for large-scale sea-ice modelling, 2024.

\end{thebibliography}

\appendix

\section{Appendix}
\subsection{Architecture and Training}
As stated in section 2.2, the building blocks of the cVAE model are convolution blocks based on a modified version of ConvNeXt blocks \cite{ConvNext2020}  as used in \cite{Samudra}. We replaced all 2D convolutions with partial convolution \cite{partialconv} layers. This is a natural choice for Arctic region where irregular land mask and small islands exist. The partial convolution layer automatically ignores these regions while processing the data. To facilitate learning using convolution operations, we break the $50\times360$ $lat\times lon$ input into two $50\times 180$ maps, reverse the second slice in latitude, and concatenate it at the top of the first slice, creating $100\times 180$ input maps that resemble a North Polar projection.

The encoder and prior networks follow the same architectures. They take the input pair ($y_{tl}$, $\bar x_{tl}$) in case of the encoder, and  ($\tilde {y}_{tl}$, $\bar x_{tl}$) in case of the prior network (we will explain below what $\tilde { y}_{tl}$ is). Additionally, we add three extra conditioning fields, uniformly inputting $\sin(\frac{2 \pi}{12} (t + l))$, $\cos(\frac{2 \pi}{12} (t + l))$, and $(\frac{l}{12})$ as input channels where $t$ is initialization time and $l$ is lead time. These networks encode their input into the mean (${\mu}_{NN_{\phi}}$ and ${\mu}_{NN_{\omega}}$)  and variance (${\sigma}_{NN_{\phi}}^2$ and ${\sigma}_{NN_{\omega}}^2$) of the latent/prior distributions. The architecture proceeds as follows:

\[\begin{aligned}
& \bullet \text{Input (5)} 
\rightarrow \text{ $3\times3$ partial convolution (16)} \rightarrow \text{ Layer normalization (16)}
\rightarrow \text{ DoubleConvNeXt (32)}\\& \rightarrow \text{ MaxPool (32)}  \rightarrow \text{ DoubleConvNeXt (64)}\rightarrow \text{ MaxPool (64)}
\rightarrow \text{ DoubleConvNeXt (128)}\rightarrow \\& \text{ MaxPool (128)} 
\rightarrow \text{ DoubleConvNeXt (256)}\rightarrow \text{ MaxPool (256)}
\rightarrow \text{ DoubleConvNeXt (256)}\rightarrow \\& \text{ Layer normalization (256)} \rightarrow \text{ Dense ($2 \times 1000$) }
\end{aligned}
\]

The decoder then takes samples from the latent space and reverses the operations in the encoder (prior) networks using upsampling and double ConvNeXt blocks. The upsampling blocks are composed of bilinear interpolation to double the resolution of the input, followed by a masked convolution with a 3 × 3 kernel smoothing the interpolated fields. The combination of interpolation with convolution results in less checkerboard effects compared to a transposed convolution \cite{Odena2016}. Finally, an output block which is a combination of layer normalization, ReLu activation and $1\times1$ convolution maps the decoder output to the SIC space. Like in Finn et al. (2024) \cite{finn2024diffice}, we use ReLu activation before the last convolution to help improve the representation of continuous-discrete sea-ice processes. The decoder proceeds as follows:

\[\begin{aligned}
& \bullet \text{Latent samples (1000)} \rightarrow \text{ Dense (256)}
\rightarrow \text{ Upsampling (256)}\rightarrow \text{ DoubleConvNeXt (128)} \\&
\rightarrow \text{ Upsampling (128)}\rightarrow \text{ DoubleConvNeXt (64)}
\rightarrow \text{ Upsampling (64)}\rightarrow \text{ DoubleConvNeXt (32)} \\&
\rightarrow \text{ Upsampling (32)}\rightarrow \text{ DoubleConvNeXt (16)} \rightarrow\\&
\text{ Layer normalization (16)} \rightarrow \text{ ReLu (16) }
\rightarrow \text{ $1\times 1$ partial convolution (1) }\end{aligned}
\]

Following Sohn et al. (2015), we create another deterministic network putting together the encoder and decoder but without the latent space in between, i.e. removing the last Layer normalization and Dense layers in the encoder, and the first Dense layer in the decoder. This network provides an initial deterministic guess of the bias corrected input ($\tilde { y}_{tl}$). This initial guess is then added as input to the prior network together with ${\bar x_{tl}}$ as mentioned above. The same output block (layer normalization + ReLu + 1$\times$1 convolution) is shared between the decoder and the deterministic network to encourage realistic outputs from the deterministic model \cite{cVAE2015}. Finally, the output of the last DoubleConvNeXt layer of the deterministic model (which has 16 channels) is summed to the corresponding layer output in the decoder before passing to the output block. 

All models were trained end-to-end using Adam optimizer, a learning rate of 0.0001 and batch size of 100. The KL divergence and log-likelihood (MSE) terms in the loss (Eq. 1) were normalized based on their dimensionalities (1000 for KL and $100\times 180$ for the log-likelihood) and the KL term was weighed with $\beta = 0.1$. The loss over the validation set was used as criterion for early stopping with a buffer of 10 epochs. The model was trained with linearly decreasing learning rate scheduler over 50 epochs after which the learning rate and scheduler were restarted and the early stopping activated after 85 epochs.

\subsection{Evaluation Metrics}

The rank of the verification data (observations) over the predicted ensemble is measured at each grid cell by sorting the ensemble members in ascending order, and ranking the observational target in that series. As stated in the text, we evaluate the rank histograms only at the marginal sea ice cells selected as grid cells with ice concentration between 0.15 and 0.9. This choice was made to avoid the results being dominated by many easy-to-predict 0 and 1 valued cells over open ocean and fully covered ice regions, respectively. For each lead time, the rankings of all marginal ice gird cell across all initialization months are pulled to plot the rank histogram. The CDF is then calculated accordingly.  

In Section 4, we also show QQ plots comparing the distributions of the corrected ensembles and observation. The QQ curve plots distribution quantiles pooling all marginal ice grid cells across all ensembles members and initialization months from the corrected ensembles (nadj and badj) at a specific lead time. This is compared to the quantiles from observations pooling all marginal ice grid cells at the same target prediction times showing how close in the distribution the corrected ensembles are to the observation.

For an ensemble of size $N$ at lead time $l$, the SOE is measured as follows \cite{SOE2013}:

\begin{equation}
    \begin{aligned}
          \quad  
          \text{SOE}_{l} = \sqrt{\frac{N+1}{N} \frac{\bar \sigma^2_{y_{tl}}}{MSE(\bar {\hat y}_{tl}, y_{tl})}}\quad
    \end{aligned}
\end{equation}

\noindent where $t$ is initialization time, $y_{tl}$ is the observation  and $\hat y_{tl}$ is the ensemble of bias adjusted forecast. $\bar{\hat y}_{tl}$ indicates ensemble mean and $\bar \sigma^2_{\hat y_{tl}}$ is variance across the $N$ ensembles averaged over initialization times. $MSE(\bar{\hat y}_{tl}, y_{tl})$ is the mean square error of the ensemble mean and observation measured over the initialization time dimension. The calculation is done at each grid cell and the area weighted average is reported at each lead time.   

For area integrated measures, the total sea ice area (SIA) refers to area integration of SIC values:

\begin{equation}
    \begin{aligned}
          \quad  
          \text{SIA}_{tl} = \int_{S_{\geq 50^{\circ}N}} SIC_{tl} \quad da\quad
    \end{aligned}
\end{equation}

\noindent where SIC could be observation ($y_{tl}$), or the ensemble mean of nadj or badj ($\bar {\hat y}_{tl}$). We report the MSE of SIA over initialization months dimension for each lead time. The same analysis is repeated for sea ice extent (SIE). SIE is a similar bulk metric as the SIA with the difference that it integrates the area of grid cells where SIC>0.15:

\begin{equation}
    \begin{aligned}
          \quad  
          \text{SIE}_{tl} = \int_{S_{\geq 50^{\circ}N}} I (SIC_{tl}>0.15) \quad da\quad
    \end{aligned}
\end{equation}

\noindent where $I(.)$ is the identity function.

Finally, we report IIEE, which captures the differences along the ice edges by quantifying the area where the predicted and true ice concentrations differ. IIEE is particularly valuable for evaluating the spatial accuracy of the ice edge locations \cite{IIEE}:

\begin{equation}
    \begin{aligned}
          \quad  
          \text{IIEE}_{tl} = \int_{S_{\geq 50^{\circ}N}} |I (y_{tl}>0.15) - I (\bar{\hat y}_{tl}>0.15)| \quad da\quad
    \end{aligned}
\end{equation}

\subsection{Scaling the standard deviation of the prior distribution }

As stated in the main text (section 2.2), the standard deviation of the prior distribution for the cVAE model was scaled at inference time based on the SOE ratio over the validation set. Although the KL divergence term in the loss function of cVAE regularizes the structure in the latent space, the samples often diverge from forming a perfect normal distribution. The cVAE benefits from the property of variational autoencoders that group similar samples closer together and structure them to the normal prior distribution through the KL term. Thus, samples belonging to more common events  are expected to be allocated to regions where the prior distribution has higher probability \cite{IEEE}, with less common samples falling on the distribution tails. The scaling factor allows sampling a wider range of internal variability at inference time \cite{IEEE,Gooya_2025}. Even without scaling, the cVAE model is still superior to the badj regarding both calibration (CDFs) and deterministic performance metrics (MSEs and IIEE). However, the resulting ensemble will be underdispersive or overconfident (Fig. A.1). Here, we show that with the proper scaling factor (3 for the validation period of $2016 - 2019$), the corrected ensemble is reliable and well-calibrated. In finding the proper scaling factor (or criterion for finding the scaling factor), caution should be taken so that widening the sampling space does not result in undercondifent ensembles, a decrease of prediction skill, or even unrealistic climate fields (hallucination). We find the SOE ratio over the validation period a reasonable metric for choosing the scaling factor while monitoring the change in other performance metrics (e.g., RMSE, QQ plots, and actual patterns) to avoid generating underconfident results, and guarantee that the generated ensemble members remain realistic. Unrealistic behavior will appear through performance metrics such as RMSE, pattern correlation, and spectral energy (analysis using spectral energy will appear in future work).


\begin{figure}[ht]
\renewcommand{\thefigure}{A\arabic{figure}}
\setcounter{figure}{0}
\centering
\includegraphics[width=\textwidth]{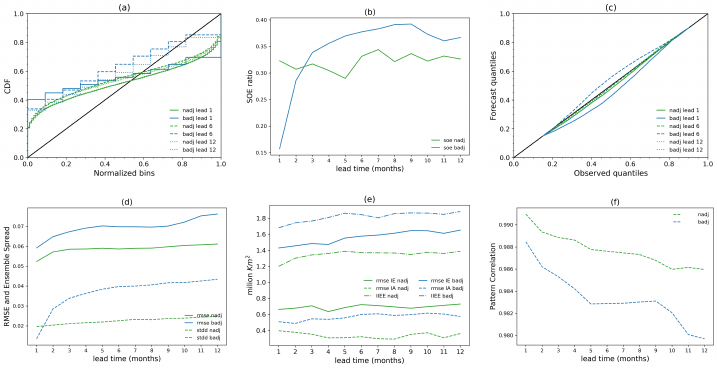}
\caption{Same as Figure 1 but for the cVAE without scaling the standard deviation of the prior distributions at inference time. a) CDF of rank histograms of the nadj/badj versus lead times measured at marginal ice grid cells. b) SOE versus lead time showing reliability. c) QQ plots at three lead times comparing the distribution of SIC at marginal ice grid cells with obs. d) RMSE (solid) over initialization time between the ensemble mean nadj/badj compared to obs at grid cells level averaged over the entire region. The dashed line shows the global mean ensemble spread averaged over initialization time. e) For each lead time, RMSE of SIA (solid line) and SIE (dashed line) over initialization time, and average IIEE (dotted line) over intialization time is compared between ensemble mean nadj/badj and obs. f) same as (e) but for pattern correlation relative to obs.} 
\label{fig:figS1}
\end{figure}



\end{document}